\documentclass{article}

\PassOptionsToPackage{numbers,compress}{natbib}

\usepackage[preprint]{neurips_2025}

\usepackage{siunitx}
\usepackage{booktabs}

\usepackage[utf8]{inputenc} 
\usepackage[T1]{fontenc}    
\usepackage{hyperref}       
\usepackage{url}            
\usepackage{booktabs}       
\usepackage{amsfonts}       
\usepackage{nicefrac}       
\usepackage{microtype}      
\usepackage{xcolor}         
\usepackage{graphicx}
\usepackage{soul}

\newcommand{\moh}[1]{\sethlcolor{yellow}\hl{[Moh: #1]}}

\newcommand{\comment}[1]{}

\newcommand{\sr}{SpeculativeReasoner}
\renewcommand{\sr}{SplitReason}

\newcommand{\srr}{SpecR}
\renewcommand{\srr}{SplitR}

\title{\sr{}: Learning To Offload Reasoning}

\author{
  \textbf{Yash Akhauri}
  \quad
  \textbf{Anthony Fei}
  \quad
  \textbf{Chi\mbox{-}Chih Chang}\vspace{0.05cm}
  \\
  \textbf{Ahmed F.\ AbouElhamayed}
  \quad
  \textbf{Yueying Li}
  \quad
  \textbf{Mohamed S.\ Abdelfattah}
  \vspace{0.1cm}\\
  \texttt{\{ya255, ayf7, cc2869, afa55, yl3469, mohamed\}@cornell.edu}\vspace{0.1cm}\\
  Cornell University\\
}

\begin{document}

\maketitle

\begin{abstract}

Reasoning in large language models (LLMs) tends to produce substantially longer token generation sequences than simpler language modeling tasks. 
This extended generation length reflects the multi-step, compositional nature of reasoning and is often correlated with higher solution accuracy.
From an efficiency perspective, longer token generation exacerbates the inherently sequential and memory-bound decoding phase of LLMs.
However, not all parts of this expensive reasoning process are equally difficult to generate.
We leverage this observation by offloading only the most challenging parts of the reasoning process to a larger, more capable model, while performing most of the generation with a smaller, more efficient model; furthermore, we \textit{teach} the smaller model to identify these difficult segments and independently trigger offloading when needed.
To enable this behavior, we annotate difficult segments across 18k reasoning traces from the OpenR1-Math-220k chain-of-thought (CoT) dataset. 
We then apply supervised fine-tuning (SFT) and reinforcement learning fine-tuning (RLFT) to a 1.5B-parameter reasoning model, training it to \textit{learn to offload} the most challenging parts of its own reasoning process to a larger model. 
This approach improves AIME24 reasoning accuracy by 24\% and 28.3\% while offloading 1.35\% and 5\% of the generated tokens respectively.
%
We open-source our \sr{} \href{https://huggingface.co/akhauriyash/DeepSeek-R1-Distill-Qwen-1.5B-GRPO-SplitReasoner}{model}, \href{https://huggingface.co/datasets/akhauriyash/OpenR1_Math_SplitReasoning}{data-set}, \href{https://github.com/abdelfattah-lab/SplitReason}{code} and \href{https://wandb.ai/akhauriyash/SplitReason}{logs}.
\end{abstract}

\begin{figure}[ht!]
    \centering
    \includegraphics[width=\linewidth]{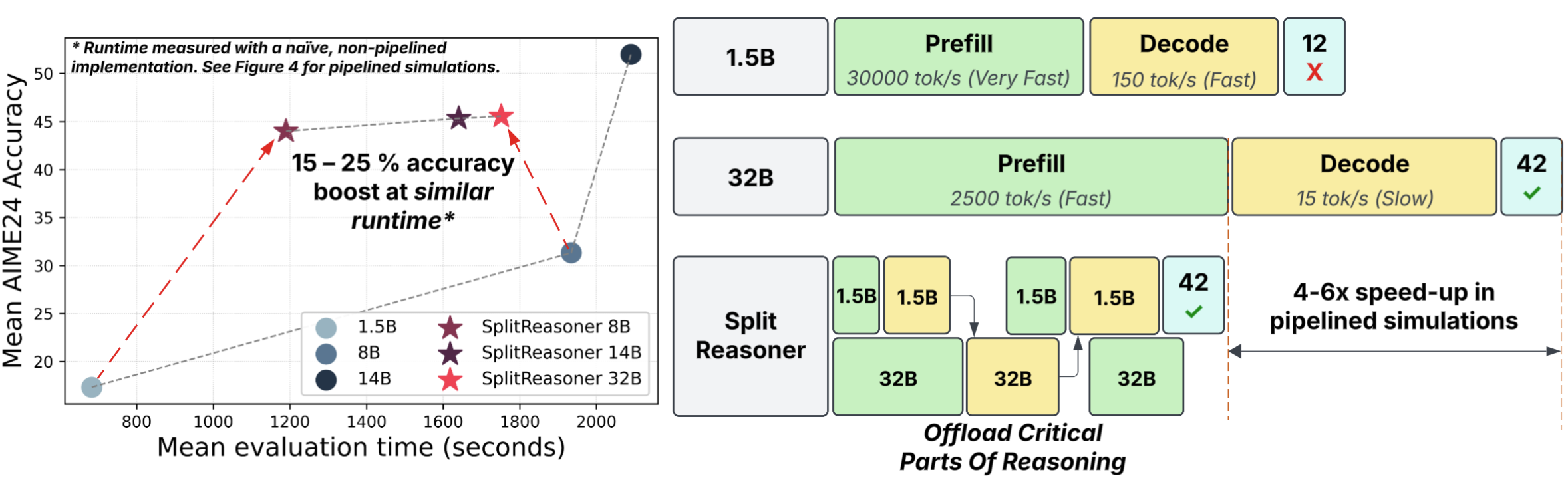}
    \caption{\sr{} intelligently offloads token generation to a large model during difficult parts of the reasoning process. Leveraging a small model (1.5B parameters) for majority of the decode process leads to significant end-to-end speedup compared to the large model (32B parameters), while improving accuracy over the small model.}
    \label{fig:teaser}
\end{figure}

\newpage
\section{Introduction}
Large language models (LLMs) are powerful general-purpose learners that excel at a wide range of tasks~\cite{brown2020languagemodelsfewshotlearners,chowdhery2022palmscalinglanguagemodeling,touvron2023llamaopenefficientfoundation}. Recent advances in LLM post-training have shown that their performance on reasoning-heavy tasks can be improved by inducing the \textit{ability to reason} by generating explicit chain-of-thoughts (CoT) about a question before arriving at the final answer \cite{NEURIPS2022_9d560961}. However, this shift towards more complex, multi-step reasoning during inference \cite{jin2024impact} significantly increases test-time compute cost. In practice, LLMs often have to generate thousands of tokens while referencing all previously produced tokens via a Key-Value Cache (KV-Cache) for every new token. This process is \textit{memory-bound} and grows \textit{quadratically} with respect to sequence length \cite{condevaux2023lsg,dao2023flashattention2fasterattentionbetter,yuan2024llminferenceunveiledsurvey}, making it very time-consuming as we scale up model sizes and rely on longer CoT to improve reasoning~\cite{wang2023selfconsistencyimproveschainthought,wei2023chainofthoughtpromptingelicitsreasoning}
In addition, further increasing compute at test time improves accuracy on reasoning tasks such as AIME24 and MATH500 \cite{muennighoff2025s1}. 
This leads to an explosion in the \textit{thinking time} needed: 
thousands of tokens are used for CoT reasoning before generating the final answer.

\begin{figure}
    \centering
    \includegraphics[width=1\linewidth]{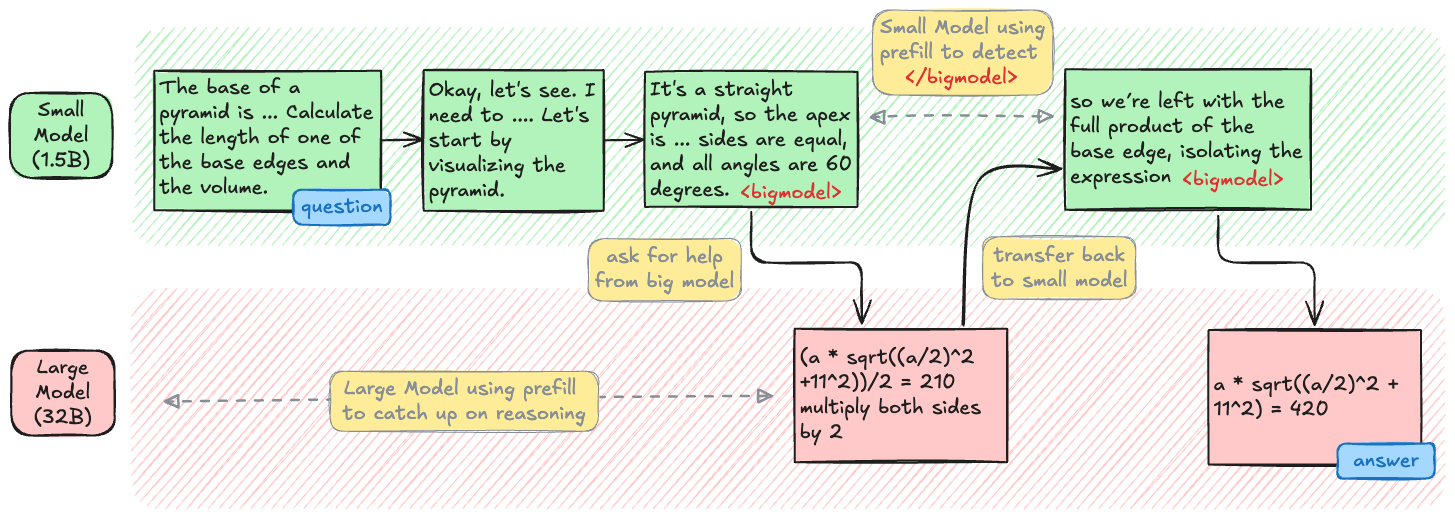}
    \caption{\sr{} utilizes two models to perform fast and high-accuracy reasoning. A small model is fine-tuned to emit a \texttt{<bigmodel>} tag when it detects a difficult reasoning step. This triggers a large model to step in and take over generation until a \texttt{</bigmodel>} tag is detected.}
    \label{fig:simple_demo}
\end{figure}

We hypothesize that reasoning segments are not uniformly difficult---certain parts of a problem can be generated with less effort using a small model, while others require more complex reasoning using larger models.
Figure~\ref{fig:simple_demo} illustrates an example of our approach: \textbf{\sr{}}. 
A small LLM (e.g., 1.5B parameters) begins generation by processing the question and starting to \textit{think} about the problem, generating an initial CoT as it reasons through the steps.
When the small model encounters a difficult part of the reasoning process, it independently emits a \texttt{<bigmodel>} token to request a reasoning segment from a much larger model (e.g., 32B parameters).
In parallel, the large model batch-processes (prefill mode) the small model's output, enabling immediate continuation of generation whenever offloading is triggered.

The roles are then reversed: the large model generates a CoT segment for the difficult part of the reasoning process (in decode mode), while the small model performs prefill on the large model’s output to immediately continue generation, or to check if the small model wants to take back control by emitting a \texttt{</bigmodel>} token.
This process can repeat multiple times until the final answer is produced.
Crucially, the most expensive part of generation---decode mode on the large model---is minimized, with the additional cost of prefill computations on both the small and large models.
A 1.5B/32B model can perform prefill and decode at $\sim$30,000/2,500 and $\sim$150/15 tokens/s respectively\footnote{Measured on A6000 GPUs on vLLM v0.8.3 for Qwen models. Two GPUs are used for models larger than 8B.}, highlighting a massive speed difference between large model decode and everything else.

%
A key challenge is determining when to switch between the small and large models based on reasoning difficulty, specifically, how to train the small model to emit the \texttt{<bigmodel>} and \texttt{</bigmodel>} tokens at the appropriate points in the reasoning process?
To achieve this, we annotate a CoT dataset (OpenR1-Math-220k) with easy and difficult segments, delimited by the special \texttt{<bigmodel>} and \texttt{</bigmodel>} tokens. 
While this annotation could be performed manually, it is far more efficient and scalable to leverage a high-quality LLM for this task; we opt for the DeepSeek-R1 671B model. 
The resulting annotated dataset is then used to perform supervised fine-tuning (SFT) on the small model, training it to insert the special tokens at appropriate points in the reasoning process. 
Finally, we apply reinforcement learning fine-tuning (RLFT) to regulate and encourage the emission of the \texttt{<bigmodel>} token, balancing downstream task accuracy against overall generation latency.
Our methodology is general and can be applied to different model efficiency approaches beyond \sr{}.
Generally, \textbf{R}einforcement \textbf{L}earning \textbf{for} optimizing \textbf{E}fficiency (\textbf{RL4E}) introduces a new paradigm by which we use fine-tuning to enable LLMs to become inherently more efficient by including measures of hardware efficiency during fine-tuning.
We enumerate our contributions below:

\begin{itemize}
    \item We develop and open-source a  fine-tuning dataset and recipe, to enable models to learn when to offload their own reasoning process to a larger model.
    \item We demonstrate that accuracy of small reasoning models can be improved by 28.3\% by offloading $\sim$5\% of the reasoning process to larger models. This can speed up inference by $4-6\times$.
    \item We show that models can learn when a task is difficult, and can leverage RL4E to attain higher efficiency. This enables a new paradigm in which models are taught to align not just with human preferences, but with hardware preferences too.
\end{itemize}

\section{Background}


\textbf{Test-Time Scaling:}
Early work on prompting showed that pretrained LLMs can reason if provided explicit CoT instructions in the prompt \cite{wei2023chainofthoughtpromptingelicitsreasoning}. However, this method is brittle and has a large inference-time token budget requirement. A more robust method to induce reasoning is with \textbf{Supervised Fine-Tuning} (SFT) on high-quality CoTs. The model is shown questions formatted with \texttt{<think>} CoT \texttt{</think>} answer, which teaches the model to imitate the reasoning trajectory. SFT has been used to \textit{induce} an internal \texttt{<think>} stage that can be exploited at test time~\cite{zelikman2022starbootstrappingreasoningreasoning}. 
However, SFT is fundamentally an imitation procedure, where the policy is rewarded for matching every token in the CoT, even if they are not decisive for getting the right answer. As a consequence, the model doesn't receive a signal to indicate that a particular step is a dead end. Reinforcement Learning (RL) tries to fill this gap, by giving rewards dependent on the \textit{outcome} (correctness) as well as rewards for formatting (for e.g., whether \texttt{<think>} tokens were used, answer returned in expected format, etc.). 
Simple outcome-level RL only look at final answers, but process-level RL \cite{kumar2024traininglanguagemodelsselfcorrect, lightman2023letsverifystepstep, gulcehre2023reinforcedselftrainingrestlanguage, zhang2024restmctsllmselftrainingprocess} also attaches rewards to intermediate steps. 
DeepSeek‑R1 introduced \textbf{Group Relative Policy Optimization (GRPO)}, a lightweight policy‑gradient variant that estimates the baseline by z‑scoring rewards \emph{within} each sampled group of trajectories, eliminating the value network and halving memory cost\cite{deepseekai2025deepseekr1incentivizingreasoningcapability, shao2024deepseekmathpushinglimitsmathematical}. 
In combination, SFT \textit{induces} the \texttt{<think>} (reasoning) behavior, while GRPO (and related RL variants) refine it. 
This two-stage recipe has given rise to several reasoning models, and motivates our own investigation in inducing tokens that can improve both accuracy \textit{and} performance.

\textbf{Performance Implications Of Test-Time Scaling:} 
Inference-time reasoning scaling strategies broadly focus on sequential and parallel scaling. Sequential approaches allocate extra compute on a single chain-of-thought, for example, by prompting the model to \textit{think longer} or iteratively refine its own output \cite{madaan2023selfrefineiterativerefinementselffeedback}. 
Such self-refinement allows LLMs to critique and improve its answer, yielding higher accuracy. Parallel approaches run multiple reasoning chains concurrently and aggregate the results, for example, by using self-consistency or best-of-N voting ~\cite{wang2023selfconsistencyimproveschainthought}. 
Sequential scaling often yields a better return on ``net tokens produced" compared to parallel scaling ~\cite{aggarwal2025l1controllinglongreasoning, muennighoff2025s1simpletesttimescaling}. However, these gains come at a significant cost; longer output means more tokens have to be decoded at inference time. 
Autoregressive generation has two distinct phases -- a \textbf{prefill} (batch processing of input tokens) and \textbf{decode} (generate tokens one-by-one). 
The prefill is a one-time, highly parallel pass over the input sequence. It has large matrix-multiplications that fully utilize the hardware's compute throughput. On the other hand, decode emits tokens one at a time; each step performing small matrix-vector operations and repeatedly fetching KV-Caches for \textit{all previous tokens}~\cite{yuan2024llminferenceunveiledsurvey}.
This makes decoding \textbf{memory bandwidth bound}, and much slower per token. The decode stage runs at a fraction of peak throughput. Pushing an LLM sequentially to produce very long chain-of-thought incurs quadratic time complexity in sequence length, which is fundamentally more expensive than parallel scaling. 

\textbf{Speculative Decoding.}
Speculative decoding~\cite{leviathan2023fast} accelerates inference by separating \emph{generation} and \emph{verification}.
A lightweight draft model first emits a short chunk of candidate tokens in the usual decode loop; a stronger verifier model then consumes the same chunk in a single, highly‑parallel pre‑fill pass. If every candidate matches the verifier’s top prediction, the entire chunk is accepted; otherwise decoding resumes from the first mismatch. The method leverages two empirical observations: (\emph{1}) even difficult language‑modeling tasks contain many locally easy continuations that a small model can approximate, and (\emph{2}) pre‑fill is markedly less memory‑bound than token‑by‑token decode on existing hardware. By letting the small model do most of the memory‑intensive decoding and having the large model perform only fast pre‑fill checks, speculative decoding yields substantial speed‑ups in practice \cite{chen2023acceleratinglargelanguagemodel,fu2024breaksequentialdependencyllm,bhendawade2025speculative,cai2024medusasimplellminference}.
A drawback, however, is that both models need to agree on the generated tokens: whenever the draft diverges from the verifier, both the verifier’s pre‑fill work and a portion of the draft’s decode work are wasted, and rolled-back to match the verifier's output.
Our approach eliminates this token‑level agreement requirement: the small model is trained to recognize \emph{apriori} which spans of a reasoning trace are likely to exceed its capability and to delegate only those spans, thereby avoiding costly verification of tokens it can already generate reliably.

\section{\sr{}}

\begin{figure}[t!]
    \centering
    \includegraphics[width=\linewidth]{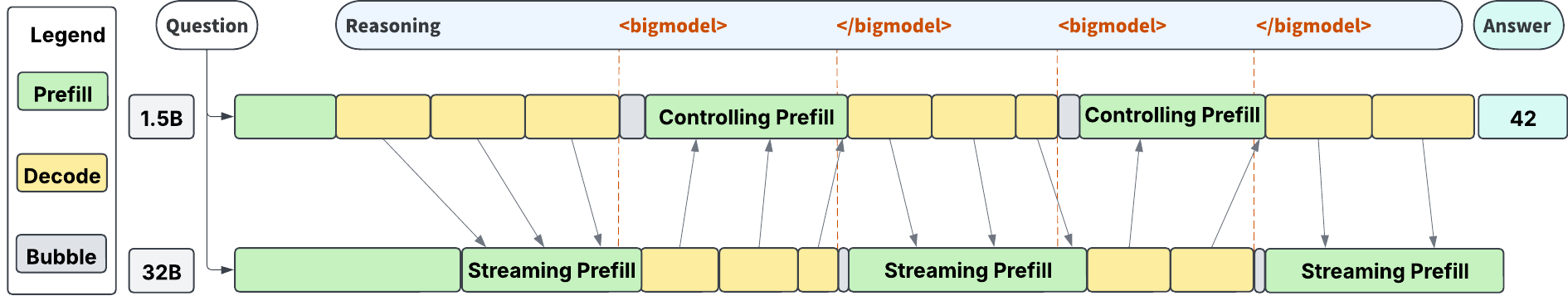}
    \caption{With \sr{}, the small model (1.5B) acts as the \textit{controller}. While the small model is decoding, the large model \textit{keeps up} with the generations by doing \textbf{streaming prefills} to keep its KV-Cache updated. Once the small model emits \texttt{<bigmodel>} tag, the large model takes over generation. At this time, the small model does \textbf{controlling prefills}, this serves a dual purpose, keeping the KV-Cache updated, as well as checking if the small model wants to take back control. The generation is halted for the large model if the small model emits \texttt{</bigmodel>} during its controlling prefill, and the small model takes over decode.}
    \label{fig:main_perf}
\end{figure}


\subsection{Cooperative Execution}
We extend the usual reasoning delimiter \texttt{<think>...</think>} with new control tokens \mbox{\texttt{<bigmodel>...</bigmodel>}}. These control tokens indicate the \textit{start and end of the offload} to the large model respectively. From Figure \ref{fig:main_perf}, the inference flow follows:
\begin{itemize}
    \item The small model is decoding. At this time, the big model does \textbf{streaming prefills}, taking chunks of small model generations and keeping its KV-Cache updated.
    \item The small model emits \texttt{<bigmodel>}, this suspends the small-model decode, and the large model starts decoding. This can happen almost immediately because the large model was performing streaming prefills to keep its KV-Cache up to-date with the current CoT trace.
    \item While the large model is generating, the small model does \textbf{controlling prefills}, taking chunks of large model generations and updating its KV-Cache, but at the same time, checking its own next-word predictions to check if it emits \texttt{</bigmodel>}, which would \textit{take back control} from the large model.
    \item Once small model emits \texttt{</bigmodel>}, the large model halts and switches to \textbf{streaming prefill}, as the small model continues the decode. 
\end{itemize}

In this flow, no modification to the large model is required. The \textbf{controlling prefill} mode continuously checks whether to halt the large model generation. Note that the prefill is highly parallel and cheap, so the small model can quickly evaluate when to halt. This method keeps the KV-Cache up to-date on both models, and either model can resume decoding without delays. Decode is memory-bound, prefill is generally compute-bound and much faster/cheaper. 
Our scheme is closely-related to speculative decoding without requiring token-by-token verification. Most ($>$95\%) of the CoT is entirely produced by the small model as we show in Section~\ref{sec:experiments}.

\begin{figure}[t!]
    \centering
    \includegraphics[width=\linewidth]{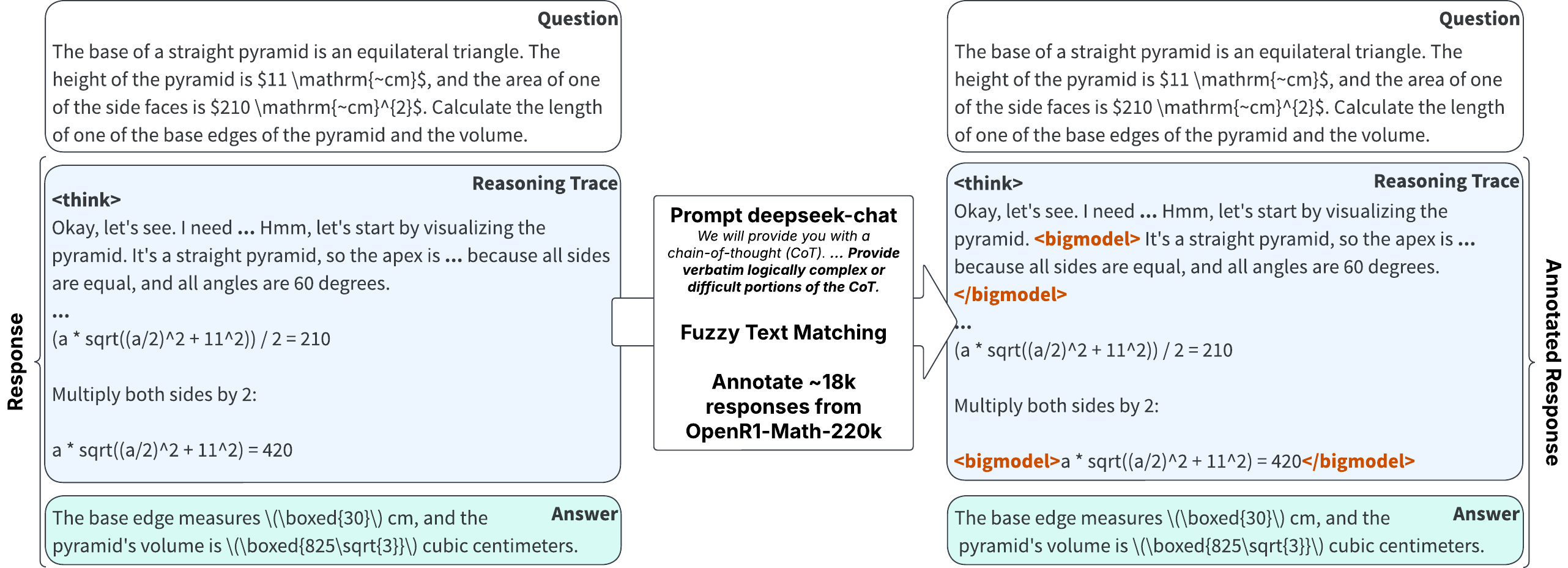}
    \caption{We take the entire response for a question from OpenR1-Math-220k and prompt deepseek-chat to annotate difficult portions of the response. These spans are encased in our \texttt{(<bigmodel>, </bigmodel>)} tags.}
    \label{fig:data_generation}
\end{figure}

\subsection{Training Procedure}
Inducing reliable offload boundaries from scratch is tricky: (\texttt{<bigmodel>...</bigmodel>}) never appear in ordinary text, so there is no reason or incentive to emit them. To address this, we follow a simple two-stage training pipeline.

\textbf{Supervised Fine-Tuning:} We sample 18k CoT traces from the Open-R1-Math-220k corpus. For each trace, we prompt deepseek-chat to annotate the most difficult spans. We then do fuzzy-text matching to identify boundaries and wrap these spans with the new control tokens (\texttt{<bigmodel>, </bigmodel>}). We take these annotations and fine-tune the small model on this corpus to induce the emission of the control tokens. We did not find it useful to make these tokens \textit{special tokens}, as the overhead of splitting these tags into tokens is negligible.

\textbf{GRPO refinement:} Supervised traces ensure the tokens appear, but they do not guarantee formatting or rewards for a target offload ratio (to control how much of the decode is offloaded, as it directly impacts latency). We therefore run GRPO on the model post-SFT using a subset of the SFT dataset. The rewards combine correctness, formatting, and \textbf{latency alignment}---a reward for adhering to the desired offload budget (e.g., 5\% of the CoT). During GRPO, we do not involve the large model for completions. This keeps the process simple, but also means that the primary focus of the reward is on latency, not on accuracy. 

This two stage pipeline is inexpensive, as it does not require the large model to be fine-tuned, and does not involve big-model invocations in the GRPO procedure. Further, \texttt{<bigmodel>} can then be offloaded to any larger model, whether it is 7B, 8B, 14B, 32B, or larger. This formulation is \textit{latency-aware}, as our offloading reward is directly calculated by simulations on expected speedup. This fine-tuning process can be further improved by real-time latency feedback and accuracy modeling with true offloading. 

\subsection{Data Generation and Training Setup}
To create our dataset, we prompt deepseek-chat to annotate the first 18k \textit{generations} from the OpenR1-Math-220k~\cite{openr1} dataset as shown in Figure~\ref{fig:data_generation}. Our prompt explicitly asks for the 20\% most logically complex or difficult portions of the CoT as snippets. We then do fuzzy text matching to ensure the text is identified correctly, and wrap that in the \texttt{<bigmodel>...</bigmodel>} tags. 

We use \texttt{DeepSeek-R1-Distill-Qwen-1.5B} for our small model. 
We first do supervised fine-tuning for 3 epochs on $8\times A6000$ GPUs with a batch size of 64, learning rate of $5e-5$, and a warmup ratio of 0.05. 
The learning rate follows a cosine decay schedule to zero. 
Following Open-R1~\cite{openr1}, we pack all SFT samples to the max sequence length of 16,384. The packed samples retain their original positional embeddings. 
We then perform GRPO fine-tuning on the resulting model.
We use 14 generations with a batch size of 128, maximum completion length of 4096 with an initial learning rate of $1e-6$. The temperature is set to 0.7, warmup ratio is 0.1, and we follow a cosine decay schedule.  
GRPO is performed on only a subset of our dataset (5000 random samples from 18k). 
We primarily rely on \texttt{DeepSeek-R1-Distill-Qwen-32B} as our large model, however, since both our SFT and GRPO training formulation do not require involvement from the large model, it is possible to use \textit{any model} as the large model.

For the GRPO procedure, we define a combined reward by summing three components, each weighted equally, to promote correctness, proper formatting and adherence to our desired offloading behavior. First, an \textit{accuracy reward} measures whether the final answer matches the ground truth; if there is a match, theres a +1 reward, else 0. Second, a \textit{format reward} checks whether the entire response follows the \texttt{<think>}...\texttt{</think>} and \texttt{<answer>}...\texttt{</answer>} scaffold, awarding +1 for correct scaffolding plus an additional +1 if all \texttt{<bigmodel>} tags are properly nested and closed. 
Finally, a \textit{tag count reward} grants partial credits for the presence of each essential tag (e.g., \texttt{<think>} and \texttt{</think>}), incentivizes well-formed \texttt{<bigmodel>} usage, and includes a coverage-based term that encourages moderate offloading. 
This coverage term is computed by measuring the fraction of tokens enclosed in \texttt{<bigmodel>} blocks and mapping it through a piecewise linear function that increases from 0 to +1 when coverage is below 0.4, then linearly decreases from +1 down to -1 as coverage approaches 1.0. 
Hence, minimal or excessive offloading is disfavored, while balanced usage is encouraged. If there is a mismatch in the number of \texttt{<bigmodel>} opening and closing tags, the reward is penalized, reflecting improper offload boundaries. These three partial rewards---accuracy, format, and tag count---are combined with equal weight into the final scalar reward for each sampled trajectory. Note that our GRPO procedure sets target offload at 0.4 (40\%), because we can always reject a \texttt{<bigmodel>} request (random-rejection), but we cannot \textit{induce} higher offloading post-finetuning. If we choose to do no offloading, we find that this SFT+GRPO procedure has no noticeable impact in the AIME24 accuracy, so we can always reject \texttt{<bigmodel>} request by trading off accuracy up to the models original baseline accuracy.


\section{Experiments}
\label{sec:experiments}

\begin{figure}[b!]
    \centering
    \includegraphics[width=\linewidth]{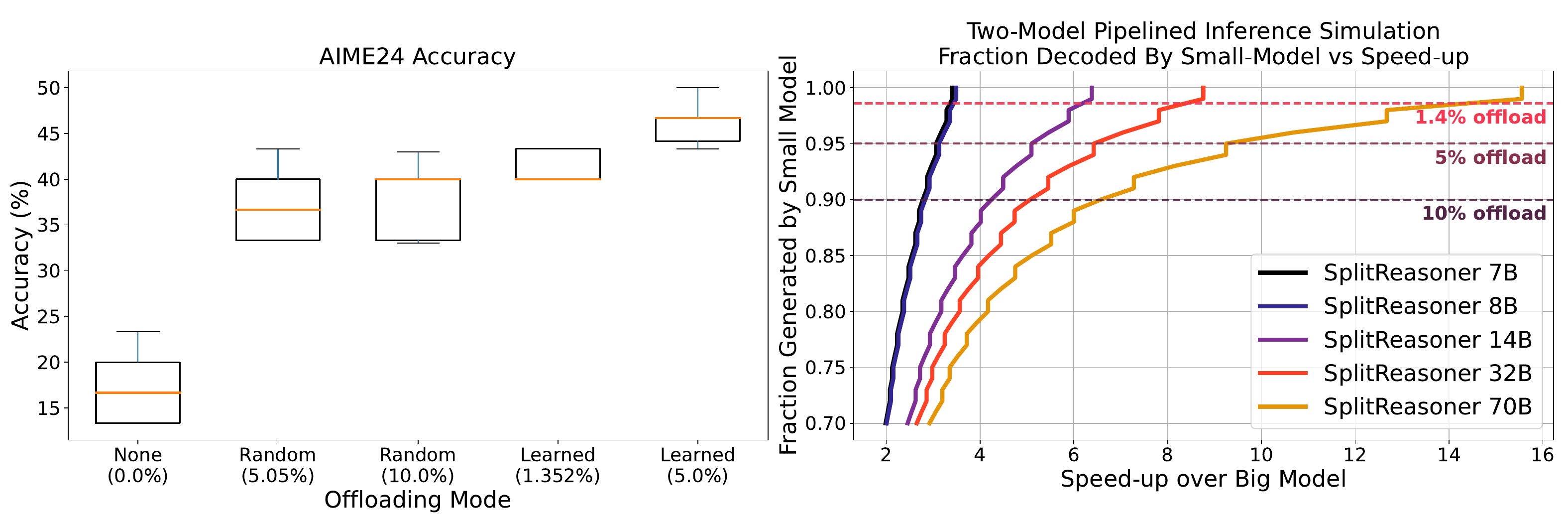}
    \caption{\textbf{(Left)} Randomly offloading sections of the decode process from a 1.5B model to 32B model boosts AIME24 accuracy by up to 20\%. Our learned offloading achieves even higher gains in accuracy (24\%--28\%) with just a 1.35\%--5\% offload. \textbf{(Right)} We run pipelined performance simulations by profiling a range of models on A6000 GPUs and find that at a 1.35\% offload, we can expect 8-9$\times$ faster inference over the large model.}
    \label{fig:random_ablation_and_inverted_scaling}
    \vspace{-6mm}
\end{figure}

\begin{figure}[h!]
    \centering
    \begin{minipage}[t]{0.49\linewidth}\vspace{0pt}
        \centering
        \includegraphics[width=\linewidth]{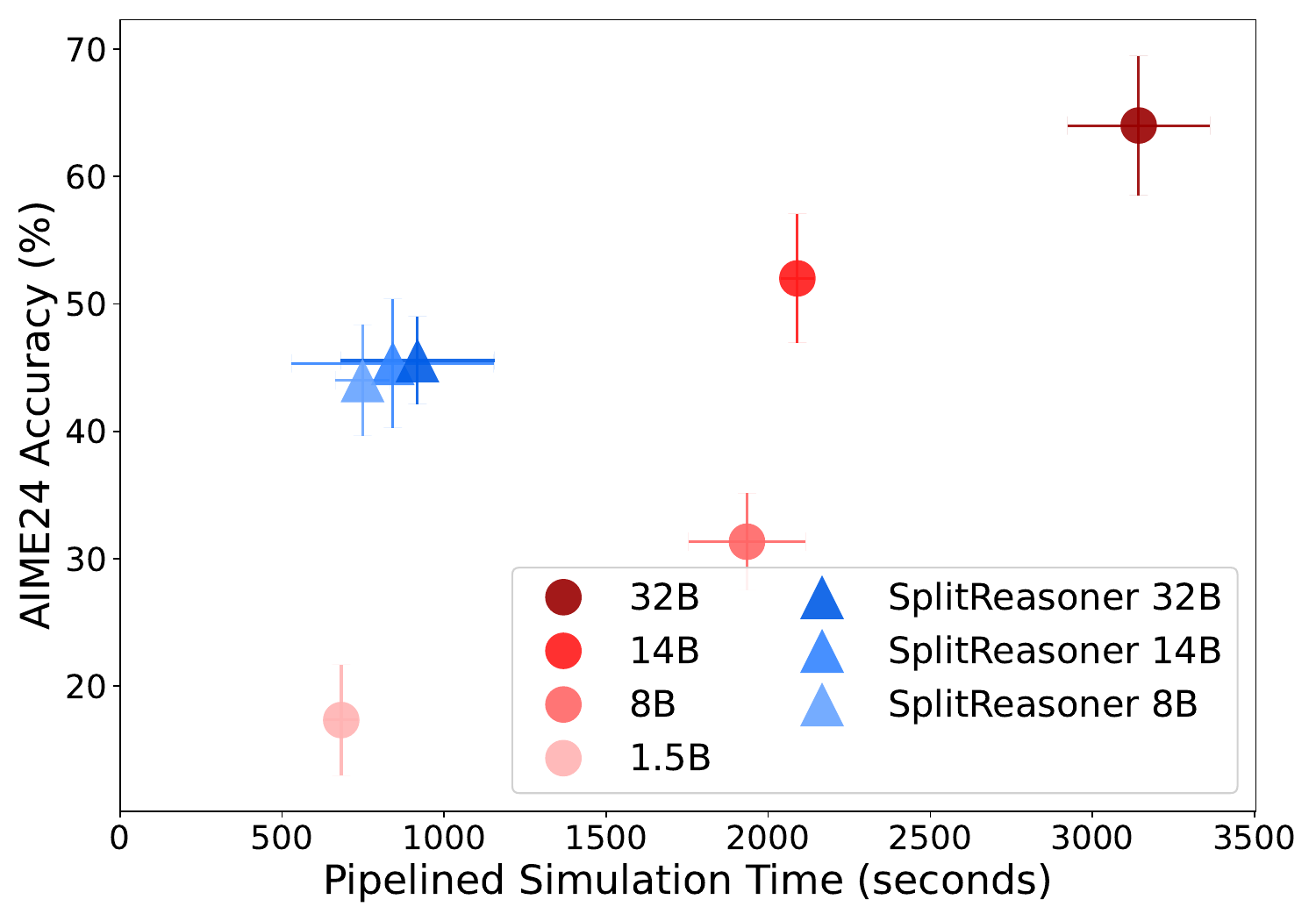}
    \end{minipage}%
    \hfill
    \begin{minipage}[t]{0.49\linewidth}\vspace{8pt}
        \centering
        \footnotesize
        \renewcommand{\arraystretch}{1.15}
        \sisetup{table-number-alignment = right}
        \begin{tabular}{
            lccc
        }
            \toprule
            {Model} & {Acc.\,(\%)} & {Non‑Pipe.\,(s)} & {Pipe.\,(s)}\\
            \midrule
            1.5B      & 17.3 & {--} &  683\\
            8B        & 31.3 & {--} & 1934\\
            14B       & 52.0 & {--} & 2090\\
            32B       & 64.0 & {--} & 3143\\
            \midrule
            \srr{}-8B  & 44.0 & 1190 &  749\\
            \srr{}-14B & 45.3 & 1640 &  842\\
            \srr{}-32B & 45.6 & 1805 &  918\\
            \bottomrule
        \end{tabular}
    \end{minipage}

    \caption{\textbf{(Left)} \sr{} (\srr{}) can benefit greatly from offloading even to smaller models: \srr{}‑8B performs almost as well as \srr{}‑14B and \srr{}‑32B while offloading only $\sim$5\% of the decode. \textbf{(Right)} \sr{} Pipe. (Pipelined) evaluation times are simulated by accounting for the 5.54\% offload overhead relative to the 1.5B baseline.}
    \label{fig:acc_lat_scatter_table}
\end{figure}

\subsection{Offloading Behavior}

Figure \ref{fig:random_ablation_and_inverted_scaling} presents the accuracy (left) and our pipelined latency simulations (right), and reveal three key observations. \textbf{(1)} Naive offloading is surprisingly effective -- randomly handing off just 5-10\% of the decode steps to the 32B model already lifts AIME24 accuracy by ~20\%. \textbf{(2)} Learning \textit{where} to offload is vastly more effective. With just a 1.35\% median offload, the small model invokes the large model only at the hardest part, exceeding accuracy of naive random offloading at 10\%. Learned offloading of 5\% of the generation pushes the accuracy further, improving by 28\% over the baseline model. \textbf{(3)} Tiny offloads translate into large simulated speed-ups, as annotated in Figure \ref{fig:random_ablation_and_inverted_scaling} (right). We find that even with a non-pipelined implementation in Figure \ref{fig:acc_lat_scatter_table} (right), it is more effective to offload to the 8B model, than to run the 8B model (1190 vs. 1934). SplitReason by design utilizes more GPUs, but the large model is less memory-bound and decodes only 5\% of the sequence. Thus, it can deliver higher through-put by serving multiple queries.

\subsection{Offloading Across Model Sizes}
\label{subsec:offload_modelsize}
One of the key advantages of \sr{} is that only the small model needs to \textit{learn to offload}, and our GRPO fine-tuning procedure does not require the larger model to be involved. To study the impact on accuracy across different \textit{large models}, we use \texttt{DeepSeek-R1-Distill-Qwen-1.5B} as the small model with \texttt{DeepSeek-R1-Distill-Llama-8B}, \texttt{DeepSeek-R1-Distill-Qwen-14B} and \texttt{DeepSeek-R1-Distill-Qwen-32B} as the large model. Figure \ref{fig:acc_lat_scatter_table} (left) reports AIME24 accuracy when the \sr{} small model delegates to large models of 8B, 14B and 32B parameters. Offloading to the smallest \textit{large} model (8B) already lifts accuracy from 17.3\% to 44\%. Accuracy continues to improve with larger models. The table in Figure \ref{fig:acc_lat_scatter_table} (right) distinguishes two runtime measurements, \emph{Non-Pipe.} corresponds to our current prototype, which executes the extra pre-fill serially after the \texttt{<bigmodel>} tag is produced. The small model further has to do several unoptimized controlling pre-fill - decode checks to verify if the small model will emit the \texttt{</bigmodel>} token. \emph{Pipe.} is a simulation of the pipelined execution presented in \ref{fig:main_perf}, with a observed offload ratio set to 5.54\%. The overhead is calculated with respect to the small model. Under this pipelined execution setting, we can see that \sr{}-32B boosts accuracy by 28\% while only marginally increasing runtime, significantly better than using a 8B/14B model. To further verify that the accuracy gains arise from offloading rather than additional fine-tuning of the small model, we re-evaluate the SFT + GRPO 1.5B checkpoint with offloading disabled; its accuracy did not improve. Thus, the improvements in Figure \ref{fig:acc_lat_scatter_table} are likely attributed to \sr{} with offloading, not a stronger small model.


\subsection{Dataset Distribution and Inducing Offloading}
\begin{figure}[t]
    \centering
    \includegraphics[width=\linewidth]{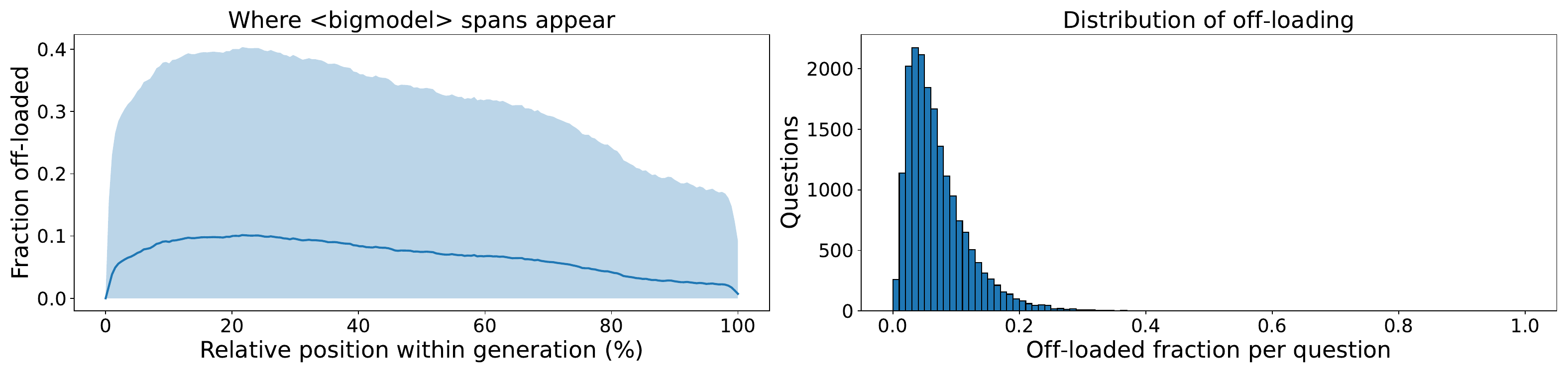}
    \caption{Analysis of our annotated dataset reveals that \textbf{(Left)} \texttt{<bigmodel>} tags appear relatively uniformly over the text, with slight preference in the earlier part of the reasoning trace and \textbf{(Right)} most questions are with-in our desired $<20\%$ offloading range.}
    \label{fig:dataset_analysis}
\end{figure}

In Figure \ref{fig:dataset_analysis}, we analyze our annotated dataset of 18,500 reasoning traces.
First, we investigate \textit{where} deepseek-chat decides to offload. Specifically, we track the relative positions of \texttt{<bigmodel>} spans across all examples, and find that there is a slightly higher bias towards offloading earlier parts of the reasoning process. This is intuitive, as the later parts of the reasoning process may just be performing compositions of prior \textit{more difficult} reasoning steps. 
Our data generation procedure adheres to the $~20\%$ offloading target, with a majority of the examples offloading less than 20\% of the trace.

\begin{figure}[h]
    \centering
    \includegraphics[width=\linewidth]{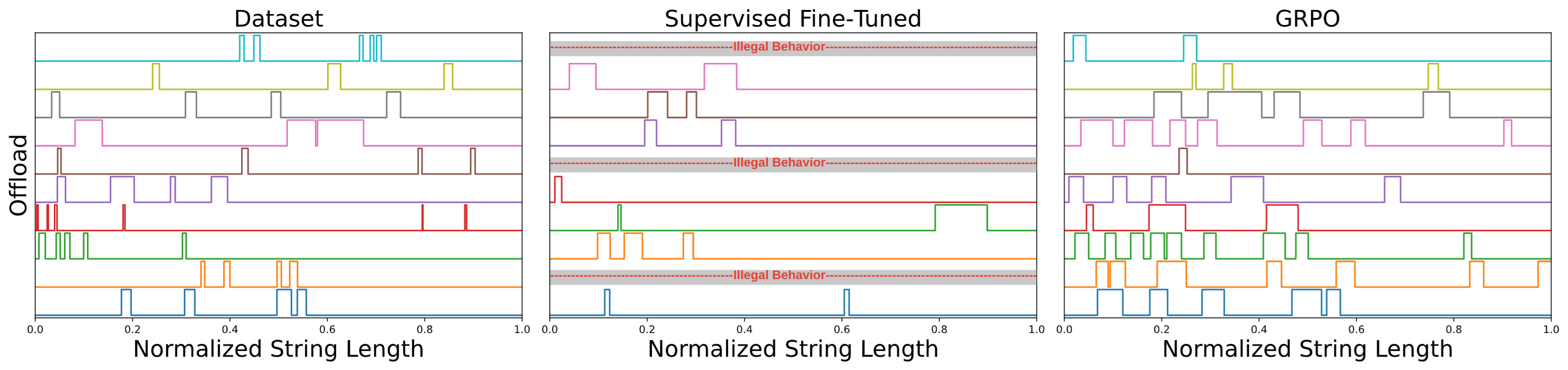}
    \caption{Stacked random samples of offloading behavior from our dataset, the post-SFT model and the final post-GRPO model. The \textit{high} signal means that part of the decode was offloaded to the large model. \textit{Illegal} offloading behavior indicates that the small model did not take back control or had incorrect formatting. The supervised fine-tuned model offloads less than 1\% of the decode and is not reliable, whereas the final model is able to reliably offload decode, adhering to our reward function.}
    \label{fig:offload_comparisons}
\end{figure}

In Figure \ref{fig:offload_comparisons}, we randomly sample 10 questions and graph the spans where the offloading occurs. 
The \textit{high signal} indicates that the span is encased in \texttt{<bigmodel>} tag. 
We find that supervised fine-tuning is not sufficient, as several generations do not have proper offloading behavior. However, after the GRPO fine-tuning, the model is able to offload effectively, following the formatting and frequency requirements. While our GRPO procedure maximizes reward for an offload of 40\%, we still empirically observed approximately a 5\% offload rate, indicating that our GRPO procedure may need further tuning.

\begin{figure}
    \centering
    \includegraphics[width=\linewidth]{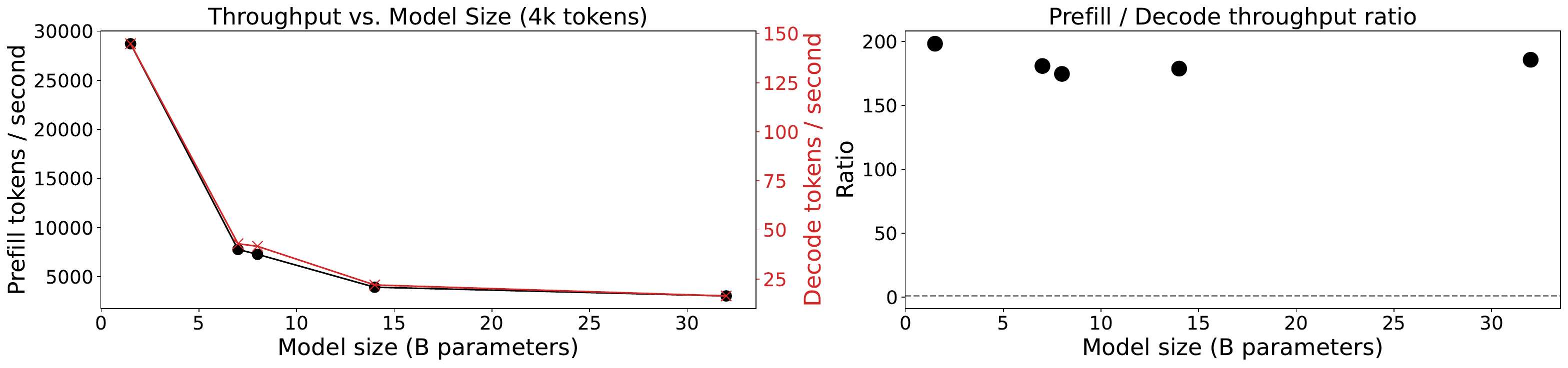}
    \caption{\textbf{(Left)} Prefill and Decode throughput decreases drastically as model size increases. Decoding \textit{most tokens} from a small model will drastically improve end-to-end latency. \textbf{(Right)} Prefill can be upto 200$\times$ faster given sufficient input sequence length, further, large model prefill is \textit{still faster} than small model decode (~3000 tokens/sec vs. 150 tokens/sec). This indicates that the large model will be able to keep up with the small model generation.}
    \label{fig:latency_plots}
\end{figure}

\subsection{Performance Simulation}
For our accuracy evaluations on AIME24, we adapt the lm-evaluation-harness~\cite{eval-harness} changes from s1 \cite{muennighoff2025s1simpletesttimescaling} with-in our own framework which uses vLLM v0.8.3 for fully-parallel evaluation of all questions efficiently. Our current implementation does not yet pipeline streaming and controlling prefills with generation. 
Our accuracy evaluation code performs prefill on the large model \textit{after} the small model generates the \texttt{<bigmodel>} token. 
Our controlling and streaming prefills operates on chunks of 64 tokens, and we constantly check the output of the controlling prefills for a \texttt{</bigmodel>} token.
%
While this naive implementation \textit{is still faster than running just the large model}, it still lacks parallelism, and incurs additional latency from the added prefill steps that can be hidden.
We simulate pipelined performance to model the concurrent small-large prefill-decode execution shown in Figure \ref{fig:main_perf}. 
Our inference simulation numbers in Figure \ref{fig:random_ablation_and_inverted_scaling} (right) are generated by profiling models of sizes \texttt{1.5B, 7B, 8B, 14B, 32B, 70B} on A6000 GPUs to feed the appropriate prefill and decode throughput numbers to our simulator. 
We present our prefill and decode throughput in Figure \ref{fig:latency_plots}. 
Given a sufficiently long input sequence, prefill is significantly faster than decode. While a small model (1.5B) is over $8\times$ faster at decode than a large model (32B) it is still slower than the large model (32B) prefill, indicating that our proposed pipelined inference flow is feasible.

\section{Discussion}

\textbf{Alignment with Latency:} In this paper, we propose to use control tokens (\texttt{<bigmodel>}) and latency-aware feedback (in the GRPO reward formulation) to demonstrate that it is possible to use RL for optimizing efficiency (\textbf{RL4E}), not just human preferences. This gives rise to several interesting questions on how to leverage control tokens to teach a model to optimize its own inference. This could be in forms beyond just offloading, such as quantization, pruning, and other compression methods. 

\textbf{Limitations:} We primarily focus on keeping an efficient training flow, this means our GRPO formulation does not actually offload the generation to the large model when the small model emits a \texttt{<bigmodel>} token.
This makes the accuracy portion of the reward unrepresentative of the actual downstream accuracy.
Instead, our current GRPO formulation only serves to encourage emitting the \texttt{<bigmodel>} tag, and to adhere to proper formatting without diverging from the original model too much.
Further, our current performance measurements are \textit{simulation based}. 
We anticipate significantly better offloading behavior may be induced by actually modeling true offloading accuracy and latency in the reward formulation. 
Further, our method still requires huge KV-Caches, as both the small and large model have to retain their KV-Cache. 
In fact, we currently also require more devices (GPUs) to run both the large and small model separately, however, since the large model has much lower device utilization, it could theoretically serve a lot more queries and is thus \textit{amortized} over batching -- beyond the large model decode, these costs are similar to that of speculative decoding and can be optimized in the same way.

\section{Conclusion}
We introduce \sr{}, a methodology by which a small model can \textit{learn to offload} reasoning to a larger model to optimize performance and accuracy. This is a novel optimization to reasoning models, where we aim to use RL for optimizing efficiency (\textbf{RL4E}), aligning language models with \textit{performance} criteria. Using the SFT+GRPO recipe to regulate \texttt{<bigmodel>} boundaries, the 1.5B model improves AIME24 accuracy by 28\%. Surprisingly, even random offloading can boost accuracy. Since most of the decoding remains on the small network, our pipelined simulation projects over \textbf{5$\times$} lower latency than running the larger, 32B model alone. The larger model itself is never fine-tuned; any model can be swapped in without re-training, demonstrating that \textbf{RL4E} can align language models to hardware objectives. 
\newpage

\bibliography{references}

\begin{thebibliography}{28}
\providecommand{\natexlab}[1]{#1}
\providecommand{\url}[1]{\texttt{#1}}
\expandafter\ifx\csname urlstyle\endcsname\relax
  \providecommand{\doi}[1]{doi: #1}\else
  \providecommand{\doi}{doi: \begingroup \urlstyle{rm}\Url}\fi

\bibitem[Brown et~al.(2020)Brown, Mann, Ryder, Subbiah, Kaplan, Dhariwal, Neelakantan, Shyam, Sastry, Askell, Agarwal, Herbert-Voss, Krueger, Henighan, Child, Ramesh, Ziegler, Wu, Winter, Hesse, Chen, Sigler, Litwin, Gray, Chess, Clark, Berner, McCandlish, Radford, Sutskever, and Amodei]{brown2020languagemodelsfewshotlearners}
Tom~B. Brown, Benjamin Mann, Nick Ryder, Melanie Subbiah, Jared Kaplan, Prafulla Dhariwal, Arvind Neelakantan, Pranav Shyam, Girish Sastry, Amanda Askell, Sandhini Agarwal, Ariel Herbert-Voss, Gretchen Krueger, Tom Henighan, Rewon Child, Aditya Ramesh, Daniel~M. Ziegler, Jeffrey Wu, Clemens Winter, Christopher Hesse, Mark Chen, Eric Sigler, Mateusz Litwin, Scott Gray, Benjamin Chess, Jack Clark, Christopher Berner, Sam McCandlish, Alec Radford, Ilya Sutskever, and Dario Amodei.
\newblock Language models are few-shot learners, 2020.
\newblock URL \url{https://arxiv.org/abs/2005.14165}.

\bibitem[Chowdhery et~al.(2022)Chowdhery, Narang, Devlin, Bosma, Mishra, Roberts, Barham, Chung, Sutton, Gehrmann, Schuh, Shi, Tsvyashchenko, Maynez, Rao, Barnes, Tay, Shazeer, Prabhakaran, Reif, Du, Hutchinson, Pope, Bradbury, Austin, Isard, Gur-Ari, Yin, Duke, Levskaya, Ghemawat, Dev, Michalewski, Garcia, Misra, Robinson, Fedus, Zhou, Ippolito, Luan, Lim, Zoph, Spiridonov, Sepassi, Dohan, Agrawal, Omernick, Dai, Pillai, Pellat, Lewkowycz, Moreira, Child, Polozov, Lee, Zhou, Wang, Saeta, Diaz, Firat, Catasta, Wei, Meier-Hellstern, Eck, Dean, Petrov, and Fiedel]{chowdhery2022palmscalinglanguagemodeling}
Aakanksha Chowdhery, Sharan Narang, Jacob Devlin, Maarten Bosma, Gaurav Mishra, Adam Roberts, Paul Barham, Hyung~Won Chung, Charles Sutton, Sebastian Gehrmann, Parker Schuh, Kensen Shi, Sasha Tsvyashchenko, Joshua Maynez, Abhishek Rao, Parker Barnes, Yi~Tay, Noam Shazeer, Vinodkumar Prabhakaran, Emily Reif, Nan Du, Ben Hutchinson, Reiner Pope, James Bradbury, Jacob Austin, Michael Isard, Guy Gur-Ari, Pengcheng Yin, Toju Duke, Anselm Levskaya, Sanjay Ghemawat, Sunipa Dev, Henryk Michalewski, Xavier Garcia, Vedant Misra, Kevin Robinson, Liam Fedus, Denny Zhou, Daphne Ippolito, David Luan, Hyeontaek Lim, Barret Zoph, Alexander Spiridonov, Ryan Sepassi, David Dohan, Shivani Agrawal, Mark Omernick, Andrew~M. Dai, Thanumalayan~Sankaranarayana Pillai, Marie Pellat, Aitor Lewkowycz, Erica Moreira, Rewon Child, Oleksandr Polozov, Katherine Lee, Zongwei Zhou, Xuezhi Wang, Brennan Saeta, Mark Diaz, Orhan Firat, Michele Catasta, Jason Wei, Kathy Meier-Hellstern, Douglas Eck, Jeff Dean, Slav Petrov, and Noah Fiedel.
\newblock Palm: Scaling language modeling with pathways, 2022.
\newblock URL \url{https://arxiv.org/abs/2204.02311}.

\bibitem[Touvron et~al.(2023)Touvron, Lavril, Izacard, Martinet, Lachaux, Lacroix, Rozière, Goyal, Hambro, Azhar, Rodriguez, Joulin, Grave, and Lample]{touvron2023llamaopenefficientfoundation}
Hugo Touvron, Thibaut Lavril, Gautier Izacard, Xavier Martinet, Marie-Anne Lachaux, Timothée Lacroix, Baptiste Rozière, Naman Goyal, Eric Hambro, Faisal Azhar, Aurelien Rodriguez, Armand Joulin, Edouard Grave, and Guillaume Lample.
\newblock Llama: Open and efficient foundation language models, 2023.
\newblock URL \url{https://arxiv.org/abs/2302.13971}.

\bibitem[Wei et~al.(2022)Wei, Wang, Schuurmans, Bosma, ichter, Xia, Chi, Le, and Zhou]{NEURIPS2022_9d560961}
Jason Wei, Xuezhi Wang, Dale Schuurmans, Maarten Bosma, brian ichter, Fei Xia, Ed~Chi, Quoc~V Le, and Denny Zhou.
\newblock Chain-of-thought prompting elicits reasoning in large language models.
\newblock In S.~Koyejo, S.~Mohamed, A.~Agarwal, D.~Belgrave, K.~Cho, and A.~Oh, editors, \emph{Advances in Neural Information Processing Systems}, volume~35, pages 24824--24837. Curran Associates, Inc., 2022.
\newblock URL \url{https://proceedings.neurips.cc/paper_files/paper/2022/file/9d5609613524ecf4f15af0f7b31abca4-Paper-Conference.pdf}.

\bibitem[Jin et~al.(2024)Jin, Yu, Shu, Zhao, Hua, Meng, Zhang, and Du]{jin2024impact}
Mingyu Jin, Qinkai Yu, Dong Shu, Haiyan Zhao, Wenyue Hua, Yanda Meng, Yongfeng Zhang, and Mengnan Du.
\newblock The impact of reasoning step length on large language models.
\newblock \emph{arXiv preprint arXiv:2401.04925}, 2024.

\bibitem[Condevaux and Harispe(2023)]{condevaux2023lsg}
Charles Condevaux and S{\'e}bastien Harispe.
\newblock Lsg attention: Extrapolation of pretrained transformers to long sequences.
\newblock In \emph{Pacific-Asia Conference on Knowledge Discovery and Data Mining}, pages 443--454. Springer, 2023.

\bibitem[Dao(2023)]{dao2023flashattention2fasterattentionbetter}
Tri Dao.
\newblock Flashattention-2: Faster attention with better parallelism and work partitioning, 2023.
\newblock URL \url{https://arxiv.org/abs/2307.08691}.

\bibitem[Yuan et~al.(2024)Yuan, Shang, Zhou, Dong, Zhou, Xue, Wu, Li, Gu, Lee, Yan, Chen, Sun, and Keutzer]{yuan2024llminferenceunveiledsurvey}
Zhihang Yuan, Yuzhang Shang, Yang Zhou, Zhen Dong, Zhe Zhou, Chenhao Xue, Bingzhe Wu, Zhikai Li, Qingyi Gu, Yong~Jae Lee, Yan Yan, Beidi Chen, Guangyu Sun, and Kurt Keutzer.
\newblock Llm inference unveiled: Survey and roofline model insights, 2024.
\newblock URL \url{https://arxiv.org/abs/2402.16363}.

\bibitem[Wang et~al.(2023)Wang, Wei, Schuurmans, Le, Chi, Narang, Chowdhery, and Zhou]{wang2023selfconsistencyimproveschainthought}
Xuezhi Wang, Jason Wei, Dale Schuurmans, Quoc Le, Ed~Chi, Sharan Narang, Aakanksha Chowdhery, and Denny Zhou.
\newblock Self-consistency improves chain of thought reasoning in language models, 2023.
\newblock URL \url{https://arxiv.org/abs/2203.11171}.

\bibitem[Wei et~al.(2023)Wei, Wang, Schuurmans, Bosma, Ichter, Xia, Chi, Le, and Zhou]{wei2023chainofthoughtpromptingelicitsreasoning}
Jason Wei, Xuezhi Wang, Dale Schuurmans, Maarten Bosma, Brian Ichter, Fei Xia, Ed~Chi, Quoc Le, and Denny Zhou.
\newblock Chain-of-thought prompting elicits reasoning in large language models, 2023.
\newblock URL \url{https://arxiv.org/abs/2201.11903}.

\bibitem[Muennighoff et~al.(2025{\natexlab{a}})Muennighoff, Yang, Shi, Li, Fei-Fei, Hajishirzi, Zettlemoyer, Liang, Cand{\`e}s, and Hashimoto]{muennighoff2025s1}
Niklas Muennighoff, Zitong Yang, Weijia Shi, Xiang~Lisa Li, Li~Fei-Fei, Hannaneh Hajishirzi, Luke Zettlemoyer, Percy Liang, Emmanuel Cand{\`e}s, and Tatsunori Hashimoto.
\newblock s1: Simple test-time scaling.
\newblock \emph{arXiv preprint arXiv:2501.19393}, 2025{\natexlab{a}}.

\bibitem[Zelikman et~al.(2022)Zelikman, Wu, Mu, and Goodman]{zelikman2022starbootstrappingreasoningreasoning}
Eric Zelikman, Yuhuai Wu, Jesse Mu, and Noah~D. Goodman.
\newblock Star: Bootstrapping reasoning with reasoning, 2022.
\newblock URL \url{https://arxiv.org/abs/2203.14465}.

\bibitem[Kumar et~al.(2024)Kumar, Zhuang, Agarwal, Su, Co-Reyes, Singh, Baumli, Iqbal, Bishop, Roelofs, Zhang, McKinney, Shrivastava, Paduraru, Tucker, Precup, Behbahani, and Faust]{kumar2024traininglanguagemodelsselfcorrect}
Aviral Kumar, Vincent Zhuang, Rishabh Agarwal, Yi~Su, John~D Co-Reyes, Avi Singh, Kate Baumli, Shariq Iqbal, Colton Bishop, Rebecca Roelofs, Lei~M Zhang, Kay McKinney, Disha Shrivastava, Cosmin Paduraru, George Tucker, Doina Precup, Feryal Behbahani, and Aleksandra Faust.
\newblock Training language models to self-correct via reinforcement learning, 2024.
\newblock URL \url{https://arxiv.org/abs/2409.12917}.

\bibitem[Lightman et~al.(2023)Lightman, Kosaraju, Burda, Edwards, Baker, Lee, Leike, Schulman, Sutskever, and Cobbe]{lightman2023letsverifystepstep}
Hunter Lightman, Vineet Kosaraju, Yura Burda, Harri Edwards, Bowen Baker, Teddy Lee, Jan Leike, John Schulman, Ilya Sutskever, and Karl Cobbe.
\newblock Let's verify step by step, 2023.
\newblock URL \url{https://arxiv.org/abs/2305.20050}.

\bibitem[Gulcehre et~al.(2023)Gulcehre, Paine, Srinivasan, Konyushkova, Weerts, Sharma, Siddhant, Ahern, Wang, Gu, Macherey, Doucet, Firat, and de~Freitas]{gulcehre2023reinforcedselftrainingrestlanguage}
Caglar Gulcehre, Tom~Le Paine, Srivatsan Srinivasan, Ksenia Konyushkova, Lotte Weerts, Abhishek Sharma, Aditya Siddhant, Alex Ahern, Miaosen Wang, Chenjie Gu, Wolfgang Macherey, Arnaud Doucet, Orhan Firat, and Nando de~Freitas.
\newblock Reinforced self-training (rest) for language modeling, 2023.
\newblock URL \url{https://arxiv.org/abs/2308.08998}.

\bibitem[Zhang et~al.(2024)Zhang, Zhoubian, Hu, Yue, Dong, and Tang]{zhang2024restmctsllmselftrainingprocess}
Dan Zhang, Sining Zhoubian, Ziniu Hu, Yisong Yue, Yuxiao Dong, and Jie Tang.
\newblock Rest-mcts*: Llm self-training via process reward guided tree search, 2024.
\newblock URL \url{https://arxiv.org/abs/2406.03816}.

\bibitem[DeepSeek-AI et~al.(2025)DeepSeek-AI, Guo, Yang, Zhang, Song, Zhang, Xu, Zhu, Ma, Wang, Bi, Zhang, Yu, Wu, Wu, Gou, Shao, Li, Gao, Liu, Xue, Wang, Wu, Feng, Lu, Zhao, Deng, Zhang, Ruan, Dai, Chen, Ji, Li, Lin, Dai, Luo, Hao, Chen, Li, Zhang, Bao, Xu, Wang, Ding, Xin, Gao, Qu, Li, Guo, Li, Wang, Chen, Yuan, Qiu, Li, Cai, Ni, Liang, Chen, Dong, Hu, Gao, Guan, Huang, Yu, Wang, Zhang, Zhao, Wang, Zhang, Xu, Xia, Zhang, Zhang, Tang, Li, Wang, Li, Tian, Huang, Zhang, Wang, Chen, Du, Ge, Zhang, Pan, Wang, Chen, Jin, Chen, Lu, Zhou, Chen, Ye, Wang, Yu, Zhou, Pan, Li, Zhou, Wu, Ye, Yun, Pei, Sun, Wang, Zeng, Zhao, Liu, Liang, Gao, Yu, Zhang, Xiao, An, Liu, Wang, Chen, Nie, Cheng, Liu, Xie, Liu, Yang, Li, Su, Lin, Li, Jin, Shen, Chen, Sun, Wang, Song, Zhou, Wang, Shan, Li, Wang, Wei, Zhang, Xu, Li, Zhao, Sun, Wang, Yu, Zhang, Shi, Xiong, He, Piao, Wang, Tan, Ma, Liu, Guo, Ou, Wang, Gong, Zou, He, Xiong, Luo, You, Liu, Zhou, Zhu, Xu, Huang, Li, Zheng, Zhu, Ma, Tang, Zha, Yan, Ren, Ren, Sha, Fu, Xu, Xie, Zhang,
  Hao, Ma, Yan, Wu, Gu, Zhu, Liu, Li, Xie, Song, Pan, Huang, Xu, Zhang, and Zhang]{deepseekai2025deepseekr1incentivizingreasoningcapability}
DeepSeek-AI, Daya Guo, Dejian Yang, Haowei Zhang, Junxiao Song, Ruoyu Zhang, Runxin Xu, Qihao Zhu, Shirong Ma, Peiyi Wang, Xiao Bi, Xiaokang Zhang, Xingkai Yu, Yu~Wu, Z.~F. Wu, Zhibin Gou, Zhihong Shao, Zhuoshu Li, Ziyi Gao, Aixin Liu, Bing Xue, Bingxuan Wang, Bochao Wu, Bei Feng, Chengda Lu, Chenggang Zhao, Chengqi Deng, Chenyu Zhang, Chong Ruan, Damai Dai, Deli Chen, Dongjie Ji, Erhang Li, Fangyun Lin, Fucong Dai, Fuli Luo, Guangbo Hao, Guanting Chen, Guowei Li, H.~Zhang, Han Bao, Hanwei Xu, Haocheng Wang, Honghui Ding, Huajian Xin, Huazuo Gao, Hui Qu, Hui Li, Jianzhong Guo, Jiashi Li, Jiawei Wang, Jingchang Chen, Jingyang Yuan, Junjie Qiu, Junlong Li, J.~L. Cai, Jiaqi Ni, Jian Liang, Jin Chen, Kai Dong, Kai Hu, Kaige Gao, Kang Guan, Kexin Huang, Kuai Yu, Lean Wang, Lecong Zhang, Liang Zhao, Litong Wang, Liyue Zhang, Lei Xu, Leyi Xia, Mingchuan Zhang, Minghua Zhang, Minghui Tang, Meng Li, Miaojun Wang, Mingming Li, Ning Tian, Panpan Huang, Peng Zhang, Qiancheng Wang, Qinyu Chen, Qiushi Du, Ruiqi Ge, Ruisong
  Zhang, Ruizhe Pan, Runji Wang, R.~J. Chen, R.~L. Jin, Ruyi Chen, Shanghao Lu, Shangyan Zhou, Shanhuang Chen, Shengfeng Ye, Shiyu Wang, Shuiping Yu, Shunfeng Zhou, Shuting Pan, S.~S. Li, Shuang Zhou, Shaoqing Wu, Shengfeng Ye, Tao Yun, Tian Pei, Tianyu Sun, T.~Wang, Wangding Zeng, Wanjia Zhao, Wen Liu, Wenfeng Liang, Wenjun Gao, Wenqin Yu, Wentao Zhang, W.~L. Xiao, Wei An, Xiaodong Liu, Xiaohan Wang, Xiaokang Chen, Xiaotao Nie, Xin Cheng, Xin Liu, Xin Xie, Xingchao Liu, Xinyu Yang, Xinyuan Li, Xuecheng Su, Xuheng Lin, X.~Q. Li, Xiangyue Jin, Xiaojin Shen, Xiaosha Chen, Xiaowen Sun, Xiaoxiang Wang, Xinnan Song, Xinyi Zhou, Xianzu Wang, Xinxia Shan, Y.~K. Li, Y.~Q. Wang, Y.~X. Wei, Yang Zhang, Yanhong Xu, Yao Li, Yao Zhao, Yaofeng Sun, Yaohui Wang, Yi~Yu, Yichao Zhang, Yifan Shi, Yiliang Xiong, Ying He, Yishi Piao, Yisong Wang, Yixuan Tan, Yiyang Ma, Yiyuan Liu, Yongqiang Guo, Yuan Ou, Yuduan Wang, Yue Gong, Yuheng Zou, Yujia He, Yunfan Xiong, Yuxiang Luo, Yuxiang You, Yuxuan Liu, Yuyang Zhou, Y.~X. Zhu,
  Yanhong Xu, Yanping Huang, Yaohui Li, Yi~Zheng, Yuchen Zhu, Yunxian Ma, Ying Tang, Yukun Zha, Yuting Yan, Z.~Z. Ren, Zehui Ren, Zhangli Sha, Zhe Fu, Zhean Xu, Zhenda Xie, Zhengyan Zhang, Zhewen Hao, Zhicheng Ma, Zhigang Yan, Zhiyu Wu, Zihui Gu, Zijia Zhu, Zijun Liu, Zilin Li, Ziwei Xie, Ziyang Song, Zizheng Pan, Zhen Huang, Zhipeng Xu, Zhongyu Zhang, and Zhen Zhang.
\newblock Deepseek-r1: Incentivizing reasoning capability in llms via reinforcement learning, 2025.
\newblock URL \url{https://arxiv.org/abs/2501.12948}.

\bibitem[Shao et~al.(2024)Shao, Wang, Zhu, Xu, Song, Bi, Zhang, Zhang, Li, Wu, and Guo]{shao2024deepseekmathpushinglimitsmathematical}
Zhihong Shao, Peiyi Wang, Qihao Zhu, Runxin Xu, Junxiao Song, Xiao Bi, Haowei Zhang, Mingchuan Zhang, Y.~K. Li, Y.~Wu, and Daya Guo.
\newblock Deepseekmath: Pushing the limits of mathematical reasoning in open language models, 2024.
\newblock URL \url{https://arxiv.org/abs/2402.03300}.

\bibitem[Madaan et~al.(2023)Madaan, Tandon, Gupta, Hallinan, Gao, Wiegreffe, Alon, Dziri, Prabhumoye, Yang, Gupta, Majumder, Hermann, Welleck, Yazdanbakhsh, and Clark]{madaan2023selfrefineiterativerefinementselffeedback}
Aman Madaan, Niket Tandon, Prakhar Gupta, Skyler Hallinan, Luyu Gao, Sarah Wiegreffe, Uri Alon, Nouha Dziri, Shrimai Prabhumoye, Yiming Yang, Shashank Gupta, Bodhisattwa~Prasad Majumder, Katherine Hermann, Sean Welleck, Amir Yazdanbakhsh, and Peter Clark.
\newblock Self-refine: Iterative refinement with self-feedback, 2023.
\newblock URL \url{https://arxiv.org/abs/2303.17651}.

\bibitem[Aggarwal and Welleck(2025)]{aggarwal2025l1controllinglongreasoning}
Pranjal Aggarwal and Sean Welleck.
\newblock L1: Controlling how long a reasoning model thinks with reinforcement learning, 2025.
\newblock URL \url{https://arxiv.org/abs/2503.04697}.

\bibitem[Muennighoff et~al.(2025{\natexlab{b}})Muennighoff, Yang, Shi, Li, Fei-Fei, Hajishirzi, Zettlemoyer, Liang, Candès, and Hashimoto]{muennighoff2025s1simpletesttimescaling}
Niklas Muennighoff, Zitong Yang, Weijia Shi, Xiang~Lisa Li, Li~Fei-Fei, Hannaneh Hajishirzi, Luke Zettlemoyer, Percy Liang, Emmanuel Candès, and Tatsunori Hashimoto.
\newblock s1: Simple test-time scaling, 2025{\natexlab{b}}.
\newblock URL \url{https://arxiv.org/abs/2501.19393}.

\bibitem[Leviathan et~al.(2023)Leviathan, Kalman, and Matias]{leviathan2023fast}
Yaniv Leviathan, Matan Kalman, and Yossi Matias.
\newblock Fast inference from transformers via speculative decoding.
\newblock In \emph{International Conference on Machine Learning}, pages 19274--19286. PMLR, 2023.

\bibitem[Chen et~al.(2023)Chen, Borgeaud, Irving, Lespiau, Sifre, and Jumper]{chen2023acceleratinglargelanguagemodel}
Charlie Chen, Sebastian Borgeaud, Geoffrey Irving, Jean-Baptiste Lespiau, Laurent Sifre, and John Jumper.
\newblock Accelerating large language model decoding with speculative sampling, 2023.
\newblock URL \url{https://arxiv.org/abs/2302.01318}.

\bibitem[Fu et~al.(2024)Fu, Bailis, Stoica, and Zhang]{fu2024breaksequentialdependencyllm}
Yichao Fu, Peter Bailis, Ion Stoica, and Hao Zhang.
\newblock Break the sequential dependency of llm inference using lookahead decoding, 2024.
\newblock URL \url{https://arxiv.org/abs/2402.02057}.

\bibitem[Bhendawade et~al.(2025)Bhendawade, Belousova, Fu, Mason, Rastegari, and Najibi]{bhendawade2025speculative}
Nikhil Bhendawade, Irina Belousova, Qichen Fu, Henry Mason, Mohammad Rastegari, and Mahyar Najibi.
\newblock Speculative streaming: Fast {LLM} inference without auxiliary models, 2025.
\newblock URL \url{https://openreview.net/forum?id=jt8wI3ZzXG}.

\bibitem[Cai et~al.(2024)Cai, Li, Geng, Peng, Lee, Chen, and Dao]{cai2024medusasimplellminference}
Tianle Cai, Yuhong Li, Zhengyang Geng, Hongwu Peng, Jason~D. Lee, Deming Chen, and Tri Dao.
\newblock Medusa: Simple llm inference acceleration framework with multiple decoding heads, 2024.
\newblock URL \url{https://arxiv.org/abs/2401.10774}.

\bibitem[Face(2025)]{openr1}
Hugging Face.
\newblock Open r1: A fully open reproduction of deepseek-r1, January 2025.
\newblock URL \url{https://github.com/huggingface/open-r1}.

\bibitem[Gao et~al.(2023)Gao, Tow, Abbasi, Biderman, Black, DiPofi, Foster, Golding, Hsu, Le~Noac'h, Li, McDonell, Muennighoff, Ociepa, Phang, Reynolds, Schoelkopf, Skowron, Sutawika, Tang, Thite, Wang, Wang, and Zou]{eval-harness}
Leo Gao, Jonathan Tow, Baber Abbasi, Stella Biderman, Sid Black, Anthony DiPofi, Charles Foster, Laurence Golding, Jeffrey Hsu, Alain Le~Noac'h, Haonan Li, Kyle McDonell, Niklas Muennighoff, Chris Ociepa, Jason Phang, Laria Reynolds, Hailey Schoelkopf, Aviya Skowron, Lintang Sutawika, Eric Tang, Anish Thite, Ben Wang, Kevin Wang, and Andy Zou.
\newblock A framework for few-shot language model evaluation, 12 2023.
\newblock URL \url{https://zenodo.org/records/10256836}.

\end{thebibliography}
\bibliographystyle{unsrtnat}

\end{document}